\documentclass[letterpaper, 10 pt, conference]{ieeeconf}  % Comment this line out if you need a4paper

\IEEEoverridecommandlockouts                              % This command is only needed if 
\usepackage{graphics} % for pdf, bitmapped graphics files
\usepackage{epsfig} % for postscript graphics files
\usepackage{mathptmx} % assumes new font selection scheme installed
\usepackage{times} % assumes new font selection scheme installed
\usepackage{amsmath} % assumes amsmath package installed
\usepackage{amssymb}  % assumes amsmath package installed
\usepackage{algorithmicx}
\usepackage{algorithm}
\usepackage{algpseudocode}
\usepackage{subcaption}
\usepackage{todonotes}

\title{\LARGE \bf
Using Deep Reinforcement Learning to Learn High-Level Policies on the ATRIAS Biped}

\author{Tianyu Li$^{1}$, Akshara Rai$^{2}$,  Hartmut Geyer$^{2}$, Christopher G. Atkeson$^{2}$% <-this % stops a space
\thanks{$^{1}$Mechanical Engineering, Carnegie Mellon University, USA,
        {\tt\small tli3@andrew.cmu.edu}}%
\thanks{$^{2}$Robotics Institute, School of Computer Science, Carnegie Mellon University, USA, {\tt\small arai@andrew.cmu.edu}, {\tt\small hgeyer@cs.cmu.edu}, {\tt\small cga@cs.cmu.edu}}%
}

\begin{document}

\maketitle
\thispagestyle{empty}
\pagestyle{empty}

%%%%%%%%%%%%%%%%%%%%%%%%%%%%%%%%%%%%%%%%%%%%%%%%%%%%%%%%%%%%%%%%%%%%%%%%%%%%%%%%
\begin{abstract}

Learning controllers for bipedal robots is a challenging problem, often requiring expert knowledge and extensive tuning of parameters that vary in different situations. Recently, deep reinforcement learning has shown promise at automatically learning controllers for complex systems in simulation. This has been followed by a push towards learning controllers that can be transferred between simulation and hardware, primarily with the use of domain randomization. However, domain randomization can make the problem of finding stable controllers even more challenging, especially for underactuated bipedal robots. In this work, we explore whether policies learned in simulation can be transferred to hardware with the use of high-fidelity simulators and structured controllers. We learn a neural network policy which is a part of a more structured controller. While the neural network is learned in simulation, the rest of the controller stays fixed, and can be tuned by the expert as needed. We show that using this approach can greatly speed up the rate of learning in simulation, as well as enable transfer of policies between simulation and hardware. We present our results on an ATRIAS robot and explore the effect of action spaces and cost functions on the rate of transfer between simulation and hardware. Our results show that structured policies can indeed be learned in simulation and implemented on hardware successfully. This has several advantages, as the structure preserves the intuitive nature of the policy, and the neural network improves the performance of the hand-designed policy. In this way, we propose a way of using neural networks to improve expert designed controllers, while maintaining ease of understanding.

% of simulation-hardware discrepancies and expensive hardware samples. Recent push in deep reinforcement learning has

\end{abstract}

%%%%%%%%%%%%%%%%%%%%%%%%%%%%%%%%%%%%%%%%%%%%%%%%%%%%%%%%%%%%%%%%%%%%%%%%%%%%%%%%
\section{INTRODUCTION}

Reinforcement learning is a continuously learning and adapting framework that can generalize to multiple robots and task scenarios. Such a framework is essential for moving towards autonomous agents that can respond to changes in their environment and learn to perform well on new tasks. This potential has sparked great interest in studying reinforcement learning approaches on complex, real world problems. Recently deep reinforcement learning (DRL) has achieved very impressive results and beyond human performance on several complex long-horizon games such as AlphaGo \cite{silver2016mastering} and even starting with no human input AlphaGoZero\cite{silver2017mastering}. Similar progress has been achieved in the domain of continuous control too, dealing with very high dimensional state and action spaces. For example, Deep Deterministic Policy Gradients (DDPG) ~\cite{lillicrap2015continuous} and Trust Region Policy Optimization  (TRPO) ~\cite{schulman2015trust} can solve several control challenges in the Mujoco simulation environment \cite{todorov2012mujoco}. \cite{heess2017emergence} show that a variant of the Proximal Policy Optimization (PPO) \cite{schulman2017proximal} can learn to control a very high degree of freedom humanoid robot in Mujoco, even in very challenging environments. Similarly, \cite{rajeswaran2017towards} use a variant of DDPG to learn to do dexterous manipulation with a high degree of freedom hand in simulation.

\begin{figure}[t]
	\centering
	\includegraphics[width=.35\textwidth]{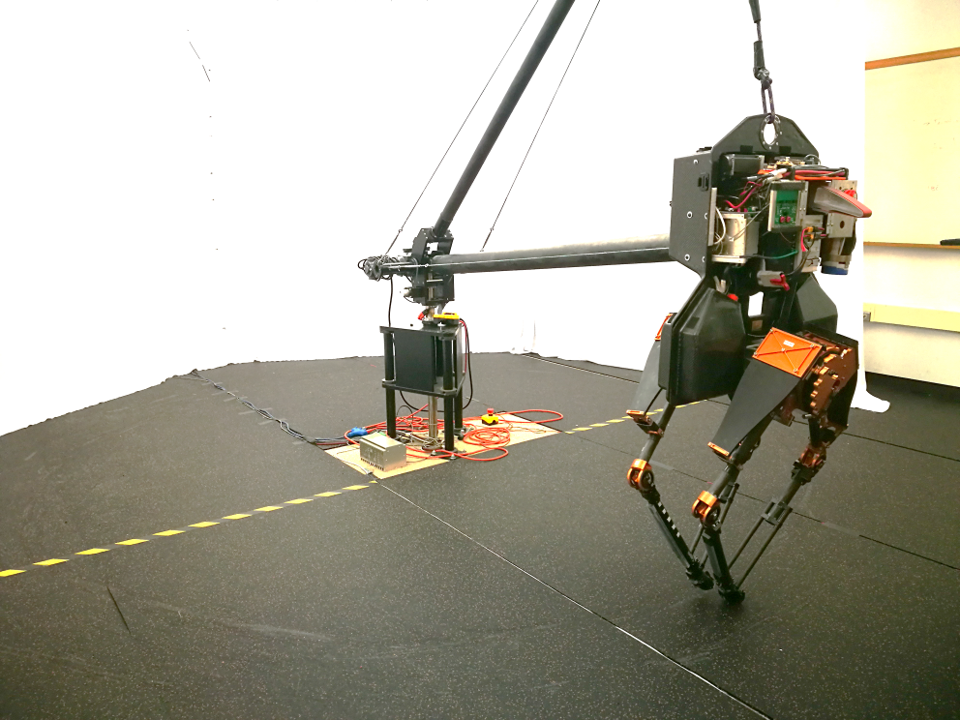}
    \caption{Our test platform is CMU's ATRIAS robot in a planar setup.}
    \label{fig:ATRIAS}
\end{figure}

% Challenge related to bipedal walking
While simulation results for deep RL are very impressive, the simulation-hardware gap makes most controllers learned in simulation unsuitable to be implemented on hardware. This is partly because simulations are not perfect representations of real systems, but also because learning approaches often tend to exploit simulation inaccuracies to achieve better performance. Recent work learns parameters of expert controllers on hardware sample-efficiently, for example \cite{antonova2017deep} and \cite{cully2015robots} use Bayesian Optimization. While these are promising, and robust even to hardware damage, they are limited by the controllers they can represent. On the other hand, high-dimensional neural network policies can approximate arbitrarily complex controllers, but too expensive to optimize on hardware. One approach to learn policies in simulation and deploy on hardware is domain randomization \cite{mordatch2015ensemble}. It can be used to learn robust neural network policies in simulation by applying random perturbations to the dynamics and other properties of the simulator. This approach has been applied to manipulation problems \cite{peng2017sim}, as well as quadrupedal locomotion problems \cite{tan2018sim}. However, typical domain randomization can lead to controllers that perform worse than controllers trained without randomization \cite{tan2018sim}, as well as make the learning problem much harder \cite{2018arXiv180800177O} taking over 100 years of simulation time to learn successful policies.   

 Domain randomization from scratch is especially challenging for under-actuated bipedal robots, as the basin of stability around a controller is typically very small. As a result, it is very difficult to randomly explore the space of controllers to find successful controllers that stabilize a wide range of dynamic models, and other disturbances. Moreover, the learned controllers would typically be quasi-static in nature, similar to \cite{mordatch2015ensemble} in behavior, which is not very impressive in performance. 
 
 In this work, we study if a high-fidelity simulator and structured neural network controllers can eliminate the need for domain randomization in learning controllers for bipedal robots. 
 %To warm start our learning process, we use behavior cloning \cite{pomerleau1991efficient} or imitation learning from an expert designed controller. Since, we can extract not just the policy but also the rewards for the expert policy, we also warm start our critic network in a supervised learning manner. Although we have a reasonable supervised initialization, further DRL training is needed to achieve good performance, especially on settings different from the expert. So we use Proximal Policy Optimization, a stable on-policy learning approach that constraints each training step to ensure that it is close to the original controller.
We use a high-fidelity simulation of the ATRIAS robot \cite{martin2015robust} with ground height disturbances to learn a walking controller in simulation. Then we test these controllers on an actual bipedal robot and evaluate if they are capable of realizing a stable walking gait on hardware. 
%To the best of our knowledge, this is the first work that demonstrates a neural network policy learned with deep reinforcement learning implemented on a bipedal robot hardware.

We experiment with different controller structures,
and study their effect on transfer from simulation to hardware, without any domain randomization. We show that structured neural network controllers have a fast training rate in simulation as well as higher rate of transfer to hardware. The structure also gives the user the power to modify the controller, if required, without having to re-train the neural network from scratch.

The main contributions of this paper are as follows:
\begin{enumerate}
    \item We demonstrate a successful neural network policy learned using deep reinforcement learning on a bipedal robot hardware
    \item We study the effect of action space on the success of transfer between simulation and hardware
    \item We present a way of using deep reinforcement learning and neural networks to improve user-designed policies, while maintaining their intuitive structure
\end{enumerate}

\section{\textsc{Learning Locomotion Controllers}}
In this section, we give a brief introduction of deep reinforcement learning and explain how we train our neural network policies in simulation.
\subsection{Background}
 We formulate our locomotion control problem as a Markov Decision Process (MDP) which can be represented as a tuple: $M = \left \{S,A,r,\tau ,T \right \}$. Here $S \subseteq \mathbb{R}^n$ is the set of states, $A \subseteq \mathbb{R}^m$ is the set of actions, $r:S \times A \to \mathbb{r}$ is the reward function, $\tau : S \times A \to S $ is the transition function which is robot's dynamics in our setting, %$\pho$ is the distribution of ground height disturbance,  
$T$ is the maximum episode length.  We wish to find a policy $\pi : S \to A$ that can maximize the expected sum of the reward over a trajectory:
\begin{equation*}
    \eta (\pi) = \mathbb{E}_{\pi} [\,\displaystyle{\sum_{i=1}^{T}} \, r_i \,]
\end{equation*}

We use an actor-critic framework to solve this problem, similar to \cite{rajeswaran2017towards}. The policy is modeled using a neural network that takes as input the partial state of the robot and outputs desired control actions. A second neural network models the value function of the robot, which is the expected total reward from the current state. %We use imitation learning to warm start our policy and then fine tune it using RL. The supervising controller is an expert controller, and the data from these demonstrations are used in to: (1) initialize the policy network, and (2) initialize the critic network using the rewards seen during demonstrations. While imitation learning can provide good a initialization, during fine tuning, this good initialization might be forgotten due to large policy update step. As a result, we use a learning method that prohibits large updates in the policy.

 %To help with exploration, we use an existing controller to generate demonstrations. We use these demonstration in two ways:(1) providing initialization for policy network; (2)calculating value function and providing initialization for critic network. Also to help with exploration, we use different high level control both based on a feedback based reactive stepping policy.While imitation learning can provide good initialization, during fine tuning using RL algorithm, we also face the problem that our good initialization might be forgotten due to large policy changing step. So we use a RL algorithm that has constraint to prohibit large policy change. In this section, we first outline the imitation learning then describe the RL method we use, for the high-level control, we will explain it in the next section.

\subsection{Imitation Learning}

Exploration is one of the challenges in RL methods; we need to try different actions in order to learn from and learn about their consequences. If we start from nothing, the learning process could be very slow \cite{rajeswaran2017learning} since the algorithms put a lot of effort on exploring state-action space. One way to reduce this issue is using behavior cloning \cite{pomerleau1991efficient}. Behavior cloning learns a policy over state-action pairs from expert trajectory \cite{ho2016generative} by finding the policy parameters $\theta$ that solve the following maximum-likelihood optimization problem \cite{rajeswaran2017learning}: 
\begin{equation}
    maximize_\theta \sum_{(s,a^*) \in \rho_D} ln \, \pi_\theta (a^*|s)
\end{equation}
where $\rho_D$ indicates the expert's trajectory distribution, with  $a^*$ is actions that expert took, and $s$ are the states visited.

Similarly, rewards accumulated during the expert's demonstrations can be used to pre-train a Q-function by minimizing TD error \cite{sendonaris2017learning}. As Q-function is the guidance for policy update, a well pre-trained Q-function benefits actor training as well. 

%Behavior cloning is one such approach that  corresponds to finding the policy parameters $\theta$ that solve the following maximum-likelihood optimization problem \cite{rajeswaran2017learning}: 
%\begin{equation}
%    maximize_\theta \sum_{(s,a^*) \in \rho_D} ln \, %\pi_\theta (a^*|s)
%\end{equation}
%where $\rho_D$ indicates the expert's trajectory distribution, with  $a^*$ is actions that expert took, and $s$ are the states visited. The target policy attempts to mimic the actions that were taken by the expert at the states visited in demonstrations. In the best case scenario, the 'student' can generate same actions as the expert, even at states not visited in the demonstrations.

In practice, imitation learning does not guarantee that the cloned policy can work as good as the expert due to compounding error caused by distributional shift in states seen during training and testing \cite{rajeswaran2017learning}, \cite{ross2010efficient}. Moreover, the expert controller might not perform sufficiently well in settings where it is difficult to design a controller. For example, while our expert controller is very stable on flat ground walking, rough ground seems to present a challenge. This means that the learned policy actually has to be updated to perform well in this situation. Nevertheless, the expert is a good initialization for the policy and makes the learning process much faster.

\subsection{Proximal Policy Optimization (PPO)}
We used PPO \cite{heess2017emergence}, \cite{schulman2017proximal} as our deep RL algorithm. PPO is a stable on-policy method that uses importance sampling and a clipped objective function to update policy parameters.  As a result, the updated policy is close to the initializing policy and the training is stable.

\section{CONTROLLERS DESCRIPTION}
In this section, we first describe a feedback based reactive policy which is also our expert controller, used for initializing our neural network policies. Then we introduce two types of neural network controllers using a similar state and action representation as the feedback based reactive policy. The first neural network controller has a general architecture, with little structure, while the second has a very similar structure to the expert.
%\subsection{ATRIAS robot} %don't know how to write unique
 %The ATRIAS robot (Figure.\ref{fig:ATRIAS}) is our test bipedal robot platform. It weights about 64kg, with its mass concentrating around its trunk, and its rotational inertia about its center of mass (CoM) is about 2.2$kgm^2$. In this work, we only focus on planar walking around a boom. Details of the robot and our simulator can be found in \cite{martin2015robust}. The simulator is designed in MATLAB Simulink environment and is very high-fidelity. It has non-linear contact models, as described in \cite{martin2015robust} for realistic contact forces, as well as detailed actuator dynamics, like spring deflection, and gear dynamics. All of this together makes the simulator an accurate representation of the robot dynamics, albeit a very slow one. However, as a result of this, we can learn controllers in simulation with hope to transfer to hardware.
 
\subsection{Feedback Based Reactive Policy (Expert Controller)}
\label{sec:expert}
This feedback based reactive policy \cite{rai2017bayesian}, is designed for controlling the CoM height and torso pitch in stance, and the leg placement location in swing:
%Our expert controller, used as a initializing controller, is a feedback based reactive stepping policy as described in \cite{rai2017bayesian}. This policy is designed for controlling the CoM height and torso pitch in stance, and the leg placement location in swing:
\begin{align}
F_x = K_{pt}(\theta_{des} - \theta) + K_{dt}(-\dot{\theta}) \\
F_z = K_{pz}(z_{des}-z) + K_{dz}( - \dot{z}) \\
x_p=k(v_{act}-v_{tgt}) + 0.5 \cdot v \cdot T 
\end{align}
Here, $F_x$ is the desired horizontal ground reaction force (GRF), $K_{pt}$ and $K_{dt}$ are the feedback gains on the torso pitch angle $\theta$ and velocity $\dot{\theta}$. $F_{z}$ is the desired vertical GRF, $K_{pz}$ and $K_{dz}$ are the feedback gains on the CoM vertical height $z$ and velocity $\dot{z}$. $\theta_{des}$ and $z_{des}$ are desired the torso lean and vertical CoM height. $x_p$ is the desired swing leg foot placement location, $v_{act}$ is the current measured horizontal velocity of CoM and $v_{tgt}$ is the target velocity. T is the swing time and the term $ 0.5 \cdot v \cdot T$ is a feedforward term similar to the Raibert controller \cite{raibert1986legged}. Our controller tries to maintain a fixed CoM height and torso pitch in stance, and regulate a forward velocity through leg placement in swing. %It assumes no double stance.

 The desired GRFs ($F_x $,$F_z$) are then sent to an inverse dynamics model of ATRIAS to generate desired motor torques that realize the GRFs \cite{wu2014highly}. These torques are tracked with low level motor velocity-based feedback loop that generates the desired current in the motors.
\begin{figure}[!h]
	\centering
	\includegraphics[width=.5\textwidth]{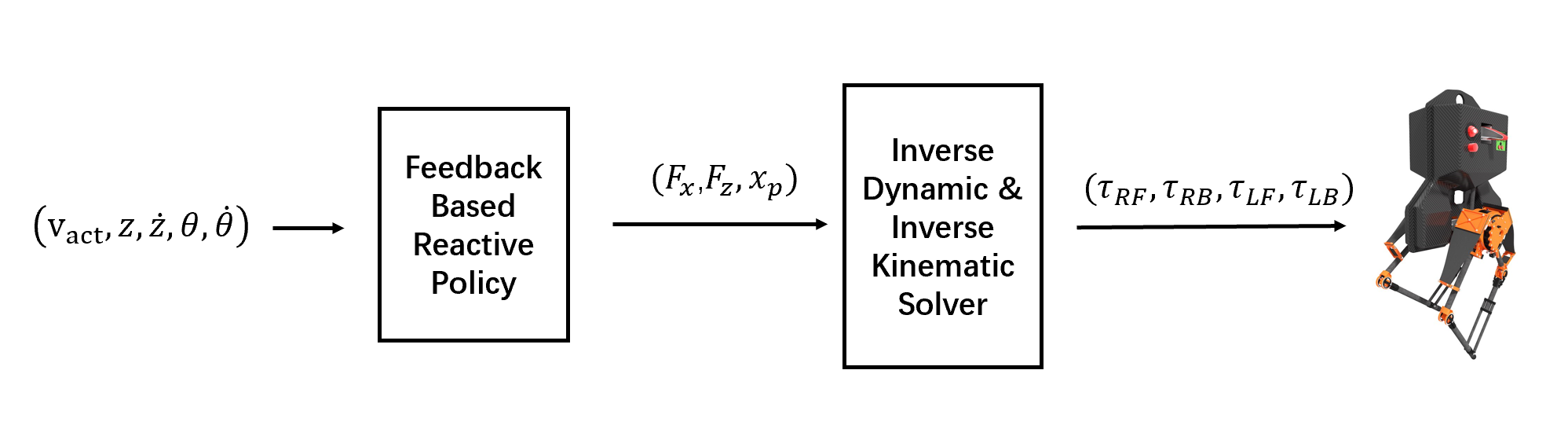}
    \caption{Pipeline of the expert controller for controlling ATRIAS.}
    \label{fig:Exp_process}
\end{figure}

The swing trajectory starts from the current leg position and terminates at the calculated foot position $x_{fp}$ with ground-speed matching. Inverse kinematics then translates this trajectory to desired joint positions and velocities. The joints are controlled by sending a velocity commands to the robot.
%\TL{Can you make a flowchart similar to the NN controllers for the expert controller?}
%This controller is based on the assumption that there is no double-stance, as soon as one leg touch the ground the other leg begin to swing. This leads to a highly dynamic gait, as the contact polygon for ATRIAS in single stance is  point. The controller also depends on the desired speed of walking which determines the desired foot landing location. This indicates that the target speed of the controller is also influential to controller's performance. The target speed is a constant value that provided by us. 

This feedback based reactive policy can walk in simulation without any perturbations. However, this controller is not robust enough, since in an environment with ground height disturbances or torque noise it often falls after a few steps. In addition, this controller cannot be directly implemented on hardware. Due to differences between the simulation and hardware such as error in models and parameters, that do not transfer, this controller is not robust enough to overcome these differences. This implies that further training is essential to achieve a better performance. We discuss two different types of controller that can be trained to obtain a better and more robust policy. 

\subsection{Neural Network Policy}
\label{NN_P}
\begin{figure}[!h]
	\centering
	\includegraphics[width=.5\textwidth]{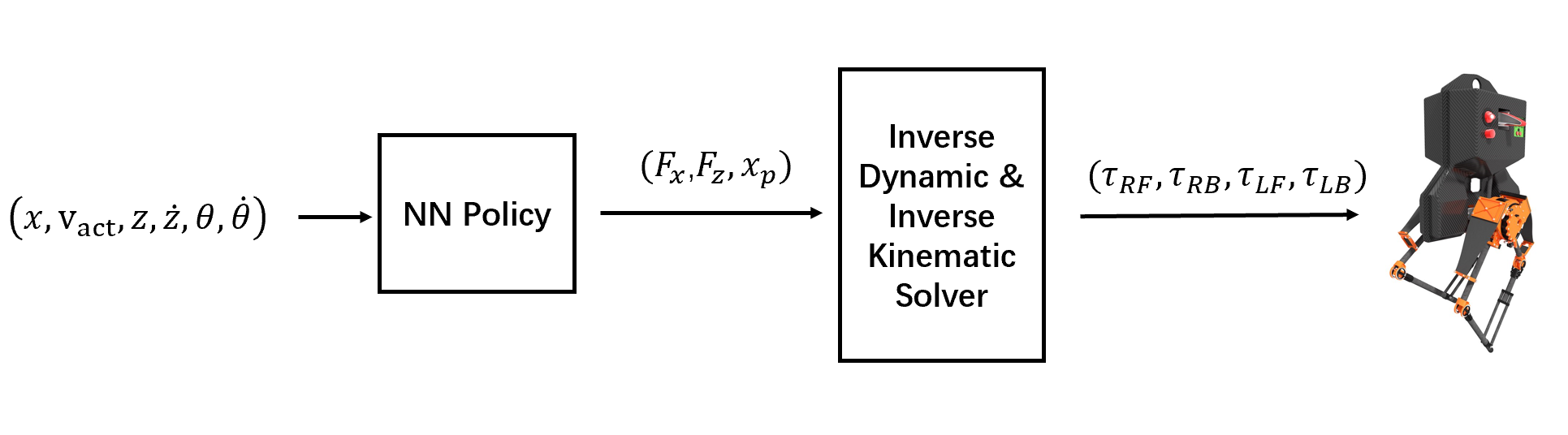}
    \caption{Pipeline of our first method for controlling ATRIAS. Here $\tau_{RF},\tau_{RB},\tau_{LF},\tau_{LB}$ are torques in the right leg's front and back and left leg's front and back motors.}
    \label{fig:NN_process}
\end{figure}

%\TL{What is the NN structure - number of layers, hidden units, activation functions?}
%\AR{I don't think most paper talked about this, I can add them on if we have enough space}
In our first setting, we create a general neural network policy without much structure. Similar to the expert controller, we design the action space of the neural network to be horizontal and vertical ground  reaction forces (GRFs) $F_z$, $F_x$ and the swing leg foot place location $x_p$. The input space is tha state of the CoM: (x,$v_{act}$,$z$,$\dot{z}$,$\theta$,$\dot{\theta}$). This means that the structure of the neural network policy becomes:
\begin{equation}
    \pi(x,v_{act},z,\dot{z},\theta,\dot{\theta})_{NN} \rightarrow  (F_z, F_x,x_p)
\end{equation}
The stance and swing control pipeline is similar to the expert and shown in Figure \ref{fig:NN_process}. %he expert controller provided above can be used to pre-train the neural network policy to walk on flat ground.

While this policy still has the information about dynamics constraints of the robot through the inverse dynamics and inverse kinematics blocks, it loses the well-defined structure of the expert. This means that the output of the neural network is unpredictable, and the user does not have a lot of control over the behavior of the robot. Roboticists typically prefer controllers that can be understood and predictable, as well as can be modified if needed. In our second controller, we try to emulate this shared autonomy, which lets the user control the overall behavior of the robot, and the neural network helps improve the user-designed policy through deep RL.

\subsection{Neural Network in the Heuristic Policy}\label{NN_HP}

In our second setting, instead of directly predicting the vertical and horizontal ground reaction force $ (F_z, F_x,x_p)$ as well as desired swing leg foot place location $x_p$, we use a neural network as part of the feedback based reactive stepping policy, described in Section \ref{sec:expert}. 
While the original policy has a fixed desired torso lean $\theta_{des}$, desired CoM height $z_{des}$ and fixed structure for $x_p$, we learn these as a function of the state of the CoM using a neural network. The neural network now takes the state of the CoM as input, and predicts the desired torso pitch, CoM height and an offset on the footstep location.
\begin{align}
  \pi(x,v_{act},z,\dot{z},\theta,\dot{\theta})_{HP} \rightarrow  (\theta_{NN}, z_{NN},x_{NN}) 
\end{align}

\begin{figure}[!h]
	\centering
	\includegraphics[width=.5\textwidth]{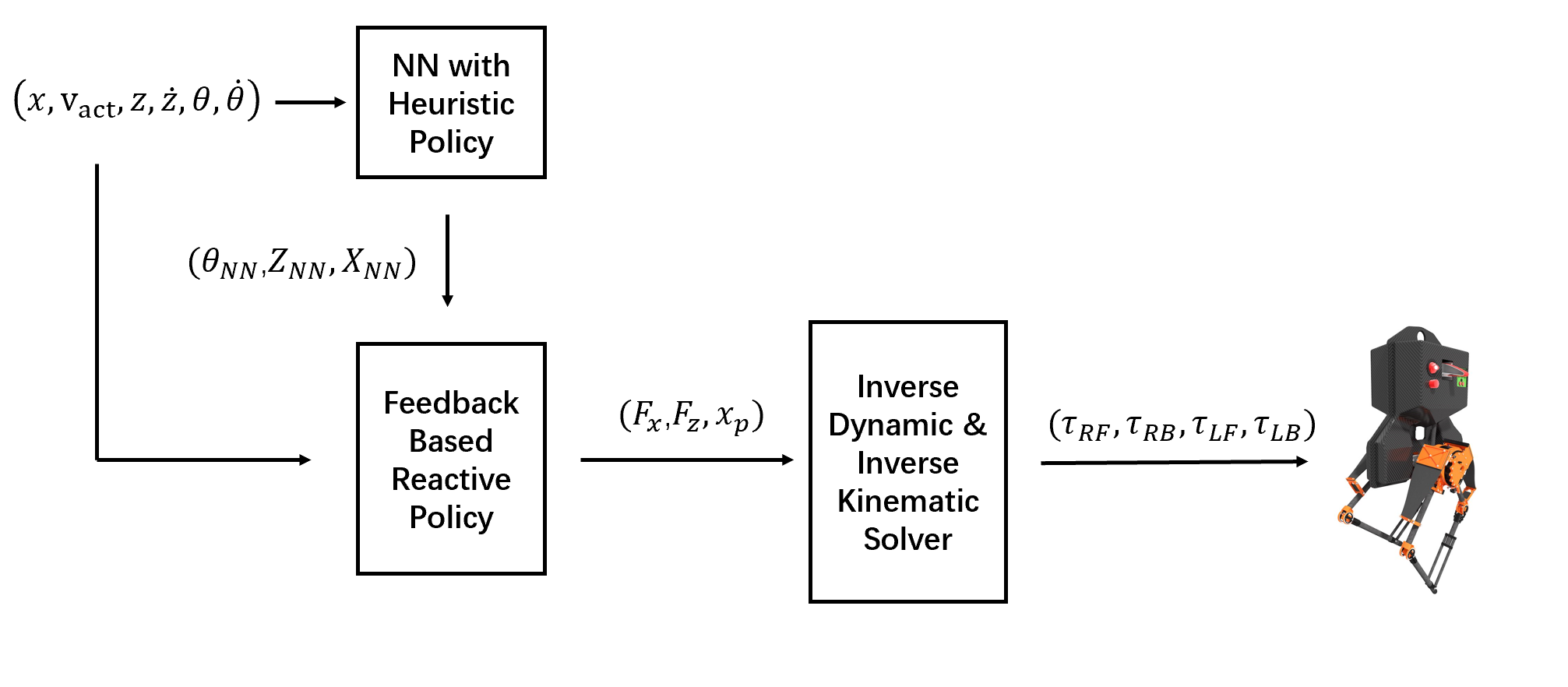}
    \caption{Pipeline of our second method using a heuristic policy and neural network for controlling ATRIAS. }
    \label{fig:H_process}
\end{figure}

When inserted into the expert controller, this gives a structured neural network policy, where neural network outputs are used as part of a heuristic policy. It is worth noting that this policy is capable of generating the same outputs as the general neural network policy. For example, for any desired $F_{x}$ profile, it is always possible to find a $\theta_{NN}$ given $K_{pt}$, $K_{dt}$, $\theta$ and $\dot{\theta}$. The pipeline is illustrated in Figure \ref{fig:H_process}.

\begin{align}
F_x = K_{pt}(\theta_{NN} - \theta) + K_{dt}(-\dot{\theta}) \\
F_z = K_{pz}(z_{NN}-z) + K_{dz}( - \dot{z}) \\
x_p=k(v_{act}-v_{tgt}) + 0.5 \cdot v \cdot T + x_{NN}\label{x_p eq}
\end{align}

%Again, the stance and swing control is similar to the expert and the pipeline is illustrated in Figure \ref{fig:H_process}.
%By doing so, we keep the structure of the expert controller while optimizing its performance. 

There are several benefits to having such structure in our policies. First and foremost, the user can predict the behavior and understand the outputs of this policy. This is important for policies implemented on robots to ensure safety of the robot and its environment. 
%For example, controllers that use on inverse dynamics take torque constraints and other dynamics constraints into account. 
Moreover, if the user wants to test a slightly different setting than the simulation, she can easily tune other parameters of this policy, keeping the neural network fixed. Lastly, if the policy fails on hardware, it is easy for the user to understand why that might be, and even possible to fix other parts of the controller without re-training the neural network.

\section{EXPERIMENT}
In this section we describe our experimental platform - the ATRIAS robot, followed by our simulation and hardware experiments. Our experiments compare a general neural network policy with a more structured policy, where the neural network helps modulate an expert controller. We compare both policies in simulation and hardware.

\subsection{ATRIAS robot}
 The ATRIAS robot (Figure.\ref{fig:ATRIAS}) is our test bipedal robot platform. It weights about 64kg, with most of its mass concentrated around its torso, and its rotational inertia about its center of mass (CoM) is about 2.2$kgm^2$. In this work, we only focus on planar walking, enforced with the help of a boom. Details of the robot and our simulator can be found in \cite{martin2015robust}. The simulator is designed in MATLAB Simulink environment and is  high-fidelity. It has non-linear contact models, as described in \cite{martin2015robust} for realistic contact forces, as well as detailed actuator dynamics, like spring deflections and gear dynamics. 
 
\subsection{SIMULATION EXPERIMENTS}
%why doing simulation
We trained our neural network controllers in simulation before implementing on hardware. This is because we use Proximal Policy Optimization (PPO) as our learning algorithm. PPO needs to collect a large amount of data points at each iteration. When our current policy is not good enough, we would not be able to collect much data since the robot might fall at the very beginning of experiment. In order to collect enough number of data points, we need to do a large number of trials, making it near impossible to train on hardware. However, since we train our policies in simulation, the resulting controllers might work well in simulation environment but when implemented on hardware, they might fall. In order to overcome this issue, we add ground height disturbances to learn robust controllers in simulation. %Furthermore, while doing training, exploration can also be regarded as one kind of extra noise added to controllers.

\begin{figure}[!h]
	\centering
	\includegraphics[width=.45\textwidth]{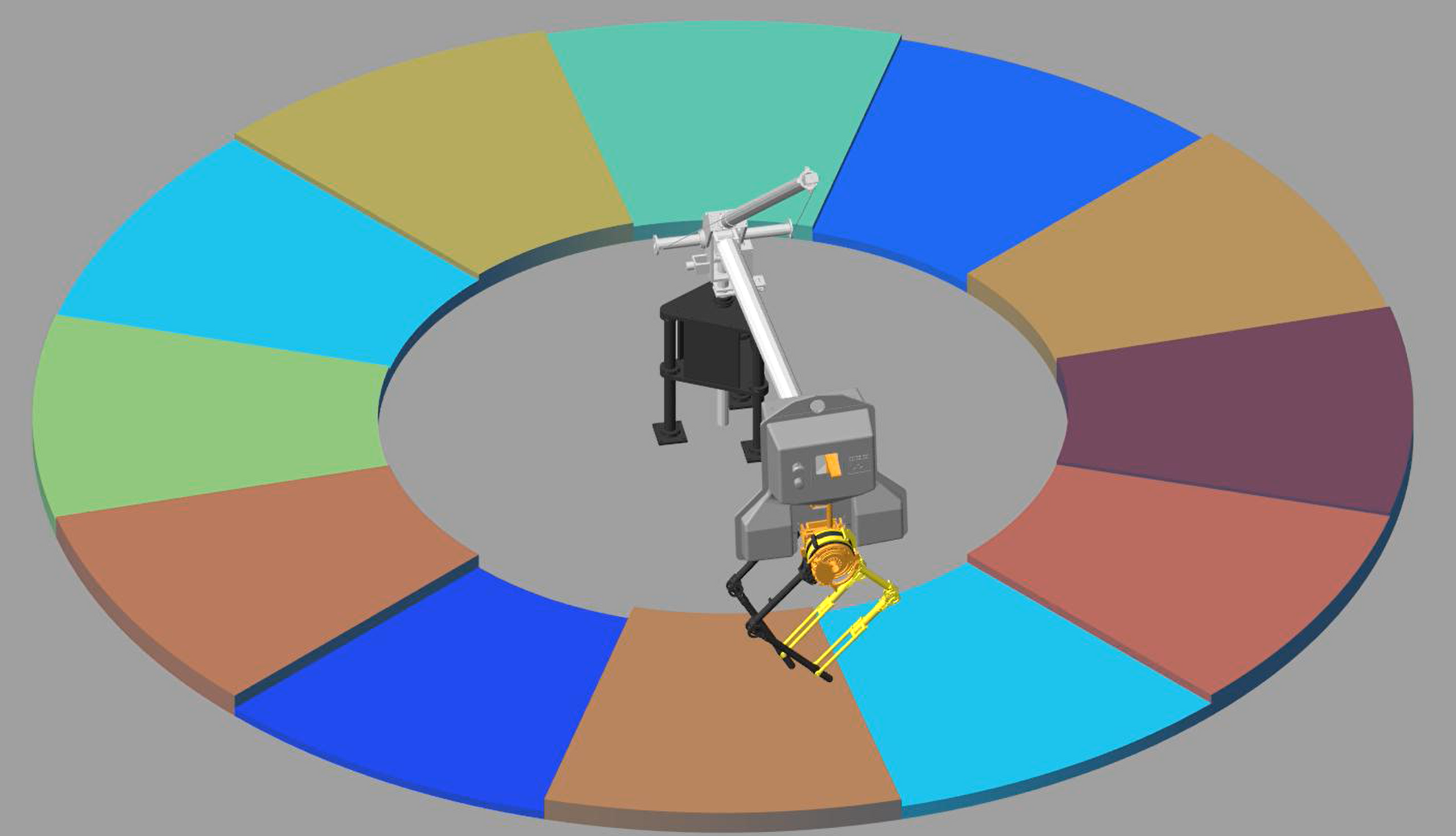}
    \caption{Simulation of ATRIAS with ground height disturbance. Different colors indicate different ground height disturbances. Maximum ground height disturbance is $\pm$ 10 cm. }
    \label{fig:process}
\end{figure}

The reward function is defined focusing on matching the desired walking speed and preventing large angular velocity of the torso. We also add a large penalty if the controller falls:
\begin{equation}
reward_t=\left\{
\begin{array}{rcl}
-C_1\cdot(v_{act, t}-v_{tgt})^2 -C_2\cdot(\dot{\theta}_t)^2 + 1 ,\  {if \ walk}\\
-C_3*T_{sim}, \  {if \ fall}
\end{array} \right.
\label{reward_1}
\end{equation}

where $v_{act, t}$ is the vector is the actual forward velocity of the robot at time $t$, $v_{tgt}$ is the target velocity, $\dot{\theta}_t$ is the angular velocity of torso pitch, $T_{sim}$ is the simulation time, $C_1,C_2,C_3$ are positive fixed parameters. This kind of cost function is similar to prior work on optimizing locomotion controllers, and it can easily distinguish points that walk from points that fall. The longer the controller can survive, the closer to the target speed, the less torso pitch the higher total reward the controller can get.
In this setting,$C_1 = 1$, $C_2=0.3$, $C_3=0.01$.
%One thing needs to be remarked is that the total reward does not directly indicate the robustness of the controller. For instance, a controller can overcome ground height disturbance might not have a high reward because it suffers from large penalty of large torso pitch which make the controller hard to be implemented on hardware. Anyway, the longer the controller can survive, the closer to the target speed, the less torso pitch the higher total reward the controller can get and more likely to work on hardware.
In our observation, often the highest scoring controllers in simulation tend to exploit simulation inaccuracies. For example, some controllers tend to jerk the torso to achieve speeds closer to the target speed. While these perform well in simulation, they tend to fail on hardware.

%\TL{Explain the set up. How many data points per iteration? 3 runs of best controller? }%we also set our data collection number up to more than 3 full simulation episodes data, so if the controller fall in simulation, we would collect more episodes of data so that we can collect enough data in one iteration.

\subsubsection{Neural Network Policy}
The first set of simulation experiments were with the Neural Network Policy (NN Policy), described in Section \ref{NN_P}. The first step for Neural Network Policy was behavior cloning. After initialization, our NN policy could walk on flat ground, but it had difficulties with ground height disturbances. This was unsurprising because the 'expert controller' also cannot walk on ground with ground height disturbance. Hence, we had to train the policy further to achieve acceptable performance.
%More importantly, the 'expert controller' has a jump up behavior at the beginning of walking, and our initialization of NN policy also inherited this behavior. In simulation, this behavior would not lead to falling, but on hardware, ATRIAS would fall as soon as it tries to start walking because of the initial jump.

\begin{figure}[!h]
	\centering
	\includegraphics[width=.45\textwidth]{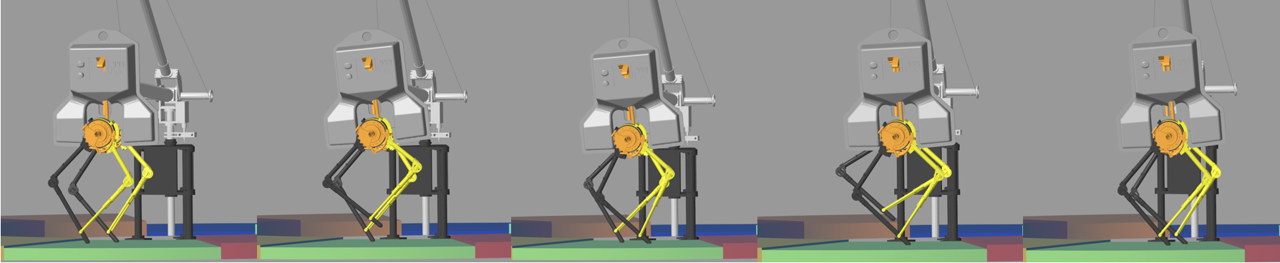}
    \caption{Sequence of frames from a single walking cycle (left to right). }
    \label{fig:NN_walking}
\end{figure}

\begin{figure}[!h]
	\centering
	\includegraphics[width=.45\textwidth]{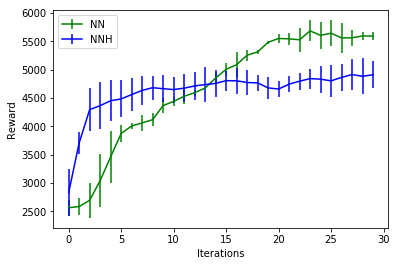}
    \caption{Training plot of Neural Network Policy (green) and Neural Network with Heuristic Policy (blue). Each of them is averaged of 5 trials data. In our training, we collected 30000 data in one iteration. For each simulation episode, we simulated 10 seconds if the controller keeps walking, and in simulation our time step was 0.001s. Thus, we could collect no more than 10000 data in a single episode, which also means in one iteration we needed data at least from 3 different episodes. By collecting data from more than one episodes, we could average the total rewards to determine the performance of the controller, also eliminating `lucky' runs.}
    \label{fig:Reward}
\end{figure}

%After initialization, we used DRL to improve the performance of our policy. 
We trained 10 different NN policies, the rewards for which are shown on Figure \ref{fig:Reward}. The rewards shown is the averaged reward of one iteration which consisted on 30,000 data points. %To reduce the influence of luck,  
As shown in the plot, the reward keeps growing as training goes on, starting from the average cost of the expert. This shows that PPO achieves a very stable training with little to no forgetting of the initial expert.

\begin{figure}[!h]
	\centering
	\includegraphics[width=.5\textwidth]{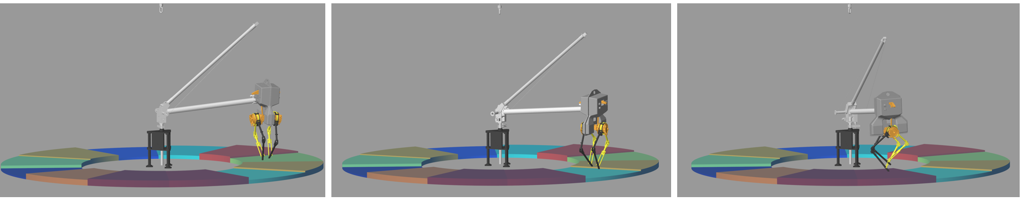}
    \caption{A time lapse of Neural Network Policy walking around the boom with ground height disturbance in simulation. }
    \label{fig:NN_walking_2}
\end{figure}

\begin{figure}[t]
	\centering
	\includegraphics[width=.45\textwidth]{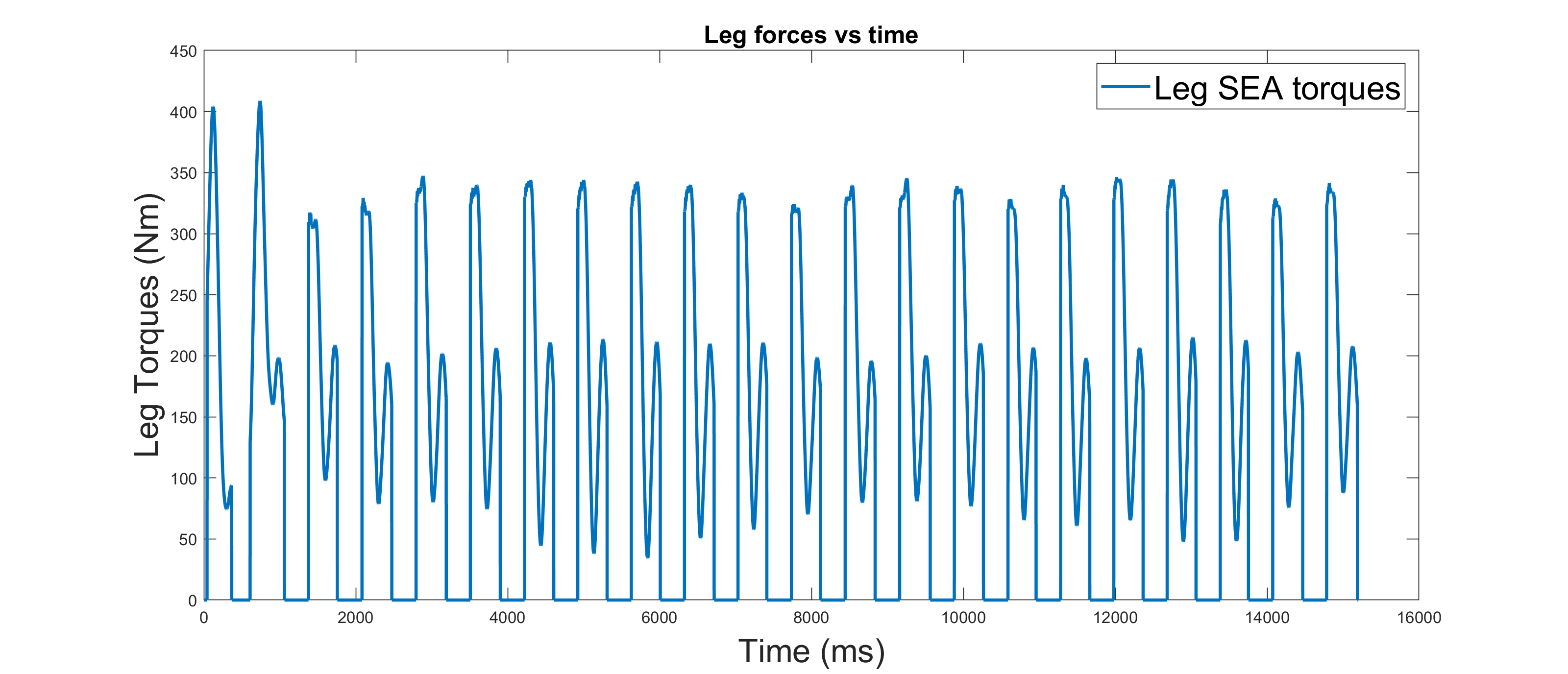}
    \caption{A plot of the stance leg SEA torques measured during one run of the NN policy on hardware. The torques are 0 during swing and follow a double hump pattern in stance. Note the high initial torques in the first two steps as compared to the rest of the walking cycle.}
    \label{fig:NN_walking_hw}
\end{figure}
The NN policy achieved good reward improvement compared to the initialization. After training, the NN policy could walk with ground height disturbance with a reasonable success rate. Figure \ref{fig:NN_walking} and Figure \ref{fig:NN_walking_2} shows the NN policy walking with ground height disturbance.

\subsubsection{Neural network with heuristic policy}

The second set of simulation experiment uses a neural network in a heuristic policy, described in section \ref{NN_HP}. %This heuristic policy don't have to do behavior cloning, we only initialized the outputs of Heuristic Policy to match the parameters of the 'expert controller'. 
By initializing this policy to the target values of the 'expert controller', the initial policy was able to walk on flat ground but cannot walk on rough ground. %And it still has a starting jump behavior.

The training plot of Heuristic Policy is shown on Figure \ref{fig:Reward}. As shown on the plot, the Heuristic Policy reward improves faster than the NN policy but reaches its peak in very few iterations. It then remains at that reward for the rest of iterations. The initial fast improvement is expected because as the heuristic policy uses the structure of the feedback-based reactive stepping policy, the search for reasonable parameters is simple. Hence, walking controllers are found much faster. However, the same structure also imposes a strong bias on the search, and as a result, it is hard to find controllers that can perform as well as the pure NN controller in simulation. However, as we describe later, even though the best reward achieved by Heuristic controllers is lower, the rate of transfer to hardware is much higher than the NN policy.

\begin{figure*}[t]
	\centering
	\includegraphics[width=.85\textwidth]{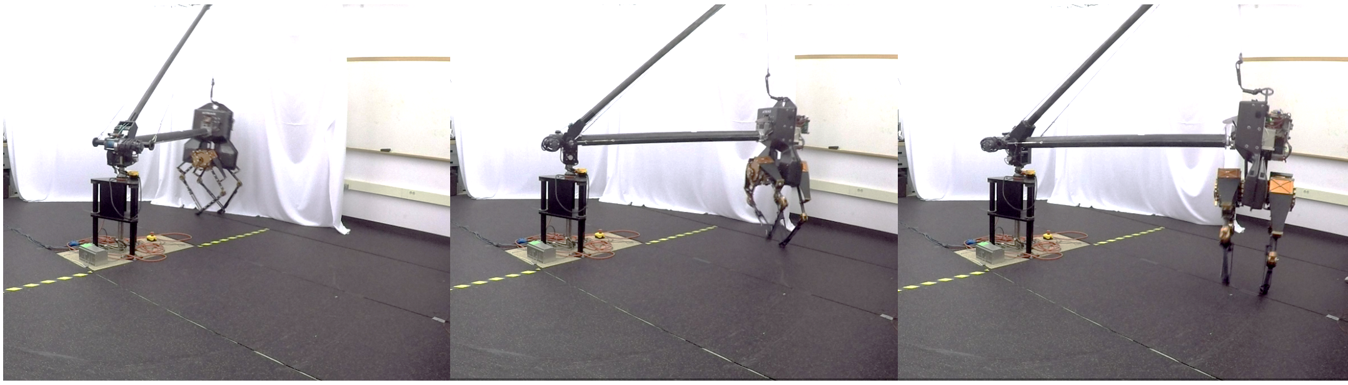}
    \caption{A time lapse of ATRIAS walking around the boom during a run of a NN with heuristic policy.}
    \label{fig:Hardware_walking}
\end{figure*}

\subsection{HARDWARE EXPERIMENTS}

We tested 5 NN policies and 5 heuristic policies in our first set of hardware experiment, with the reward function described in Function \ref{reward_1}. 

There is a large mismatch between simulation and hardware starting conditions for our system. Starting from rest is a significant challenge for bipedal robots, especially underactuated robots like ATRIAS. ATRIAS is incapable of balancing by standing in place, and needs to continue stepping to stay upright. Typically, a start-up routine is designed for these robots to reach some initial walking gait, and then the controller to be tested is initiated. For example \cite{hubicki2016atrias} start by executing a stepping motion in air, and a human holds on to the robot as its lowered onto the ground. We train our controllers to start from rest in both simulation and hardware, to avoid having to design a start-up routine. However, our simulation leg is not initially loaded, so the simulation has to make an initial push to keep the CoM from going too low. On the other hand, on hardware the robot starts in position control in the air and then is lowered. So as stance starts, there is already sufficient force in the leg, and the initial push (from simulation) makes the robot leave ground. Such a controller falls on hardware. This behavior is illustrated in Figure \ref{fig:NN_walking_hw} where the first two steps on hardware experience higher leg forces than the rest of the steps.

\subsubsection{Neural network policy}

2 out of 5 of the NN policy could walk on hardware. But all of the NN policies shared the same issue that at the beginning of walking they would jump up. This behavior can also be seen in our simulation. The initializing 'expert controller' has a jump up behavior at the beginning of walking, and our NN policy also inherits this behavior. In simulation, this behavior would not lead to falling, but on hardware, ATRIAS would fall as soon as it tries to start walking because of the initial jump. On hardware, 2 out of 5 controllers could overcome the initial disturbance and then started normal walking pattern. 3 out 5 controllers, however could not recover from this disturbance. This led to a \textbf{$40\%$} rate of transfer of learned policies between simulation and hardware.

\subsubsection{Neural network with heuristic policy}

4 out 5 controllers trained with neural network as part of the heuristic structure were able to walk on hardware. There are a few reasons for the success of the NN with heuristic structure policy, over the pure NN policy. Firstly, the initial discrepancy between simulation and hardware leg force is already compensated for by the heuristic structure. In the simulation, the NN predicts a desired height, and the feedback on this desired height results in a high leg force, as the actual height of the robot is lower. On hardware, the actual height of the robot is close to this desired height, so it automatically reduces the initial leg force. Unlike pure NN policy, where the forces were high on both simulation and hardware, now the initial forces are only high when the actual state of the robot is different from the desired. This structure in our policy, hence, makes the generalization from simulation to hardware very simple and intuitive.

Secondly, since the structure of the policy is readable by an expert, the expert can still edit the policy after being trained in simulation. For example, if the NN was trained on a lower velocity but the expert wanted to run the robot on a higher velocity, the heuristic structure allows to directly edit this target. Similarly, we observed that some of our controllers were falling because of speeding, and we increased the feedback gain $k$ on velocity feedback, described in Equation \ref{x_p eq}. This modification significantly improved the success rate of policies learned in simulation on hardware.

These experiments highlight the advantages of structure in learned policies. While the NN helped us improve our heuristic policy, as shown in Figure \ref{fig:Reward}, by learning a state-dependant desired height, pitch and footstep location, it also kept the intuitive structure of the heuristic. So, it was able to better reject disturbances between simulation and hardware, as well as be modified for slightly different test situations, without needing re-learning. This allowed a \textbf{$80\%$} rate of transfer of learned policies between simulation and hardware.

\section{CONCLUSIONS AND DISCUSSION}

In this work, we used deep reinforcement learning to learn two neural network policies to control the ATRIAS biped in simulation, and study their effectiveness at transferring to hardware. One of the policies uses a general neural network, while the second builds on the structure of a heuristic policy. We show that introducing structure into neural network policies can vastly improve the transfer rate of policies between simulation and hardware. A structured policy can guarantee the safety of the robot, as well as, allow a user to intervene if the controller fails. In our experiments, such a controller led to a $80\%$ rate of transfer between simulation and hardware, as compared to $20\%$ for a traditional NN policy. These results show that incorporating neural networks into heuristic policies can help improve the performance of the policy, and increase the rate of transfer of NN policies between simulation and hardware.

In our hardware setup, the initial condition in simulation and hardware was very different, which led to a lot of controllers failing on hardware. This is a common problem in locomotion systems, especially bipeds. One way of learning more robust policies to such disturbances can be to use cost functions that encourage conservative controllers. We conducted such experiments with a cost that penalized high torques and found that the transfer improved to a $100\%$ for NN policy with heuristic structure. In general, it is worth studying control frameworks that can switch between conservative and high performing controllers, based on the mismatch between simulation and hardware. For example, we might want to start with a conservative controller, and then transition to a high-performance controller once we have achieved a cyclic gait. We leave this for future work.

%In this work, we used DRL methods to train two types of Neural Network Policies: Neural Network Policy and Neural Network with Heuristic Policy in high-fidelity simulator and implemented these policies on real bipedal robot: ATRAIS. Using DRL methods these policies can work on simulator with ground height disturbance successfully. We find that incorporating neural network policy with a more structured controller, we can speed up the training in simulation comparing to less structured neural network policy. Although both types of policies can work in simulator, we find that more structured neural network policy is easier to transfer from simulation to hardware, since it keeps the intuitive structure. Beside that, the structured neural network policy has the advantage that it can be tuned by expert if there is any needs.

\bibliographystyle{IEEEtran}

\bibliography{references}
\end{document}